%% file: acl2012.tex
\documentclass[11pt,letterpaper]{article}
\usepackage[letterpaper]{geometry}
\usepackage{acl2012}
\usepackage{times}
\usepackage{latexsym}
\usepackage{amsmath}
\usepackage{multirow}
\usepackage{url}
\makeatletter
\newcommand{\@BIBLABEL}{\@emptybiblabel}
\newcommand{\@emptybiblabel}[1]{}
\makeatother
\usepackage[hidelinks]{hyperref}

\usepackage{graphicx}
\usepackage{subfigure}
\usepackage{arydshln}
\usepackage{amssymb,amsfonts}
\def\figref#1{Figure~\ref{fig:#1}}
\def\figlabel#1{\label{fig:#1}\label{p:#1}}

\def\tabref#1{Table~\ref{tab:#1}}
\def\tablabel#1{\label{tab:#1}\label{p:#1}}

\def\secref#1{Section~\ref{sec:#1}}
\def\seclabel#1{\label{sec:#1}\label{p:#1}}
\def\eqref#1{Eq.~\ref{eqn:#1}}

\long\def\eat#1{\ignorespaces}

\def\explainindexi{($i\in\{0,1\}$)}
\def\colfigfactor{0.95}

\title{ABCNN: Attention-Based Convolutional Neural Network\\ for Modeling Sentence Pairs}

\author{Wenpeng Yin, Hinrich Sch\"utze\\
Center for Information and Language Processing\\
LMU Munich, Germany\\
{\tt wenpeng@cis.lmu.de}\\\And
Bing Xiang, Bowen Zhou\\
IBM Watson\\
Yorktown Heights, NY, USA\\
{\tt {bingxia,zhou}@us.ibm.com}\\}

\date{}

\newcounter{notecounter}
\newcommand{\enotesoff}{\long\gdef\enote##1##2{}}
\newcommand{\enoteson}{\long\gdef\enote##1##2{{
\stepcounter{notecounter}
\large\bf
\hspace{1cm}\arabic{notecounter} $<<<$ ##1: ##2
$>>>$\hspace{1cm}}}}
\enoteson
\enotesoff

\begin{document}

\maketitle
\begin{abstract}
How to model a pair of sentences is a critical issue in many
NLP tasks such as answer
selection (AS), paraphrase identification (PI) and textual
entailment (TE). Most prior work (i) deals with one
individual task by fine-tuning a specific system; (ii)
models each sentence's representation separately, rarely considering the
impact of the other sentence; or (iii) relies fully on manually
designed, task-specific linguistic features. This work
presents a general $\textbf{A}$ttention $\textbf{B}$ased
$\textbf{C}$onvolutional $\textbf{N}$eural
$\textbf{N}$etwork (ABCNN) for modeling a pair of
sentences. We make three contributions. (i) The ABCNN can be
applied to a wide variety of tasks that require modeling of
sentence pairs. (ii) We propose three attention schemes that
integrate mutual influence between sentences into
CNNs; thus, the representation of each
sentence takes into consideration its counterpart. These
interdependent sentence pair representations are more
powerful than isolated sentence representations.  (iii)
ABCNNs achieve state-of-the-art performance on AS, PI and TE
tasks. We release code at: \url{https://github.com/yinwenpeng/Answer_Selection}.
\end{abstract}

\section{Introduction}

How to model a pair of sentences is a critical issue in many
NLP tasks 
such as answer
selection (AS) \cite{yu2014deep,feng2015applying},
paraphrase identification (PI) \cite{madnani2012re,yinnaacl},
textual
entailment (TE) \cite{marelli2014semeval,bowman2015large} etc.

Most prior work derives each sentence's representation separately, rarely
considering the impact of the other sentence.  This neglects
the mutual influence of the two sentences in the context of
the task. It also contradicts what humans do when comparing
two sentences. We usually focus on key parts of one
sentence by extracting parts from the other sentence that
are related by identity, synonymy, antonymy and other relations.
Thus, human beings model the two sentences together, using
the content of one sentence to guide the representation of
the other.

\setlength{\belowcaptionskip}{-15pt}
\setlength{\abovecaptionskip}{0pt}
\setlength{\tabcolsep}{0.5mm}
\begin{figure}
\footnotesize
\begin{tabular}{lll}
\multirow{3}{*}{\rotatebox{90}{AS}}&$\mathbf{s_0}$& how much did Waterboy \emph{gross}?\\
&$\mathbf{s}_1^+$& the movie \emph{earned} \$161.5 million\\
&$\mathbf{s}_1^-$& this was Jerry Reed's final film appearance\\\hline
\multirow{3}{*}{\rotatebox{90}{PI}}&$\mathbf{s_0}$ & she
struck a deal with RH  to pen a book \emph{today} \\
&$\mathbf{s}_1^+$ & she signed a contract with RH  to write a book\\
&$\mathbf{s}_1^-$ & she denied \emph{today} that she struck  a deal with RH\\\hline
\multirow{3}{*}{\rotatebox{90}{TE}}&$\mathbf{s_0}$& an ice skating rink placed \emph{outdoors} is \emph{full of people}\\
&$\mathbf{s}_1^+$& a \emph{lot of people} are in an ice skating park\\
&$\mathbf{s}_1^-$& an ice skating rink placed \emph{indoors} is \emph{full of people}
\end{tabular}
\caption{Positive 
($<\!s_0,s_1^+\!>$)
and negative
($<\!s_0,s_1^-\!>$)
examples for AS, PI and TE
tasks. RH = Random House\label{fig:example}}
\end{figure}

\figref{example} demonstrates that each sentence of a pair
partially determines which parts of the other sentence we
must focus on.  For AS, correctly answering $s_0$ requires
attention on ``gross'': $s_1^+$ contains a
corresponding unit (``earned'') while $s_1^-$ does not.
For PI, focus should be removed from ``today'' to correctly
recognize $<\!s_0,s_1^+\!>$ as paraphrases and
$<\!s_0,s_1^-\!>$ as non-paraphrases.  For TE, we need to
focus on ``full of people'' (to recognize TE for
$<\!s_0,s_1^+\!>$) and on ``outdoors'' / ``indoors'' (to
recognize non-TE for $<\!s_0,s_1^-\!>$).  These examples
show the need for an architecture that computes different
representations of $s_i$ for different $s_{1-i}$ \explainindexi.

Convolutional Neural Networks (CNNs) \cite{lecun1998gradient}
are widely used to model sentences
\cite{kalchbrenner2014convolutional,kim2014convolutional}
and sentence pairs
\cite{socher2011dynamic,yinnaacl},  especially
in classification tasks. CNNs are supposed to be good at
extracting robust and abstract features of input.
This work presents
the ABCNN, an attention-based convolutional neural network, that
has a powerful mechanism for modeling a sentence pair by
taking into account the interdependence between the two
sentences.  The ABCNN is a general architecture that can handle
a wide variety of sentence pair modeling tasks.

Some prior work proposes simple mechanisms that can be
interpreted as controlling varying attention; e.g.,
\newcite{yih2013question} employ word alignment to match
related parts of the two sentences. In contrast, our
attention scheme based on CNNs  models relatedness
between two parts fully automatically. Moreover, attention
at multiple levels of granularity, not only at  word
level, is achieved as we stack multiple convolution layers
that increase abstraction. 

Prior work on attention in deep
learning (DL) mostly addresses long short-term memory
networks (LSTMs)
\cite{hochreiter1997long}. LSTMs achieve attention
usually in a word-to-word scheme, and word representations
mostly encode the \emph{whole context} within the sentence
\cite{bahdanau2015neural,entail2016}. It is not clear
whether this is the best strategy; e.g., in the AS example
in \figref{example}, it is possible to determine that ``how
much'' in $s_0$ matches ``\$161.5 million'' in $s_1$ without
taking the entire sentence contexts into
account. This observation was also investigated by
\newcite{yao2013automatic} where an information retrieval
system retrieves sentences with tokens labeled as DATE by
named entity recognition or as CD by POS tagging
if there is a ``when'' question. However, labels or POS tags
require extra tools.  CNNs benefit from incorporating
attention into representations of \emph{local phrases}
detected by filters; 
in contrast, LSTMs
encode the \emph{whole context} to
form attention-based word representations -- a strategy that
is more complex than the CNN strategy and (as our
experiments suggest) performs less well for some tasks.

Apart from these differences, it is clear that attention 
has as much potential
for CNNs as it does for LSTMs. As far as we know, this
is the first NLP paper that incorporates attention into
CNNs. Our ABCNNs get state-of-the-art in AS and TE tasks,
and competitive performance in PI, then obtains further
improvements over all three tasks when linguistic
features are used.

\eat{

Section \ref{sec:relatedwork} discusses related
work. Section \ref{sec:bcnn} introduces BCNN, a network that
models two sentences in parallel with shared weights, but
without attention.  \secref{abcnn} presents three different
attention mechanisms and their realization in ABCNNs, an
architecture that is based on BCNNs.  Section
\ref{sec:experiments} evaluates the models on AS, PI and TE
tasks and conducts visual analysis for our attention
mechanism. Section \ref{sec:sum} summarizes the
contributions of this work.

}

\section{Related Work}\label{sec:relatedwork}

\input{new1}

\textbf{DL on Sentence Pair Modeling.} 
To address some of the challenges of non-DL work,
much recent work uses neural networks 
to model sentence pairs for AS, PI and TE.

For AS, \newcite{yu2014deep} present a bigram CNN to model
question and answer candidates. 
\newcite{yang2015wikiqa} extend this method 
and get state-of-the-art performance on the
WikiQA dataset (\secref{as}). 
\newcite{feng2015applying} test various setups of a
bi-CNN architecture on an insurance domain QA dataset.
\newcite{tan2015lstm} explore
bidirectional LSTMs on the same
dataset. Our approach is different because we do not model
the sentences by two independent
neural networks in parallel, but instead as an
interdependent sentence pair, using attention.

For PI, \newcite{blacoe2012comparison} form sentence
representations by summing up word
embeddings. \newcite{socher2011dynamic} use recursive
autoencoders (RAEs) to model representations of local phrases
in sentences, then pool similarity values of phrases from
the two sentences as features for binary classification.
\newcite{yinnaacl} similarly replace an RAE with a CNN.
In all three papers, the representation of
one sentence
is not influenced by the other -- in contrast
to our attention-based model.

For TE, \newcite{bowman2015recursive} use recursive
neural networks to encode entailment on SICK
\cite{marelli2014sick}.  \newcite{entail2016}
present an attention-based LSTM for the Stanford natural
language inference corpus \cite{bowman2015large}. Our system
is the first CNN-based work on TE.

Some prior work aims to solve a general sentence matching
problem.  \newcite{hu2014convolutional} present
two CNN architectures, ARC-I and ARC-II, for sentence
matching. ARC-I focuses on sentence representation learning
while ARC-II focuses on matching features on phrase
level. Both systems were tested on PI, sentence completion
(SC) and
tweet-response matching.  \newcite{yin2015ACL}
propose the  MultiGranCNN architecture to model
general sentence matching based on phrase matching on
multiple levels of granularity and get promising results for PI and
SC.  \newcite{wan2015deep} try to match two sentences in AS
and SC by multiple sentence representations, each coming
from the local representations of two LSTMs.  Our work is
the first one to investigate attention for the general
sentence matching task.

\textbf{Attention-Based DL in Non-NLP Domains.} Even though
there is little if any work on
attention mechanisms in CNNs for NLP,
attention-based CNNs have been used in
computer vision for 
visual question answering \cite{chen2015abc}, image
classification \cite{xiao2015application}, caption
generation \cite{xu2015show}, image segmentation
\cite{hong2015learning} and object localization \cite{cao2015look}.

\input{new2}
\newcite{gregor2015draw} combine a spatial attention
mechanism with RNNs for image
generation. \newcite{ba2015multiple} investigate
attention-based RNNs for recognizing
multiple objects in
images. \newcite{chorowski2014end} and \newcite{chorowski2015attention}
use attention in RNNs for speech
recognition.

\textbf{Attention-Based DL in NLP.} Attention-based DL
systems have been applied to  NLP after their success
in computer vision and speech recognition. They mainly rely
on RNNs and end-to-end encoder-decoders
for tasks such as machine translation
\cite{bahdanau2015neural,luong2015effective} and text
reconstruction \cite{li2015hierarchical,rush2015neural}. Our work takes the lead in exploring attention mechanisms in CNNs for NLP tasks.

\section{BCNN: Basic Bi-CNN}\label{sec:bcnn}
We now introduce our basic (non-attention)
CNN that is based on the Siamese architecture \cite{bromley1993signature}, i.e., it consists of two weight-sharing CNNs, each
processing one of the two sentences, and a final layer that
solves the sentence pair task.  See \figref{arc-0}.  We
refer to this architecture as the \emph{BCNN}. The next section
will then introduce the ABCNN,
an attention architecture that extends
the BCNN.  \tabref{notation} gives our notational conventions.

In our implementation and also in the mathematical
formalization of the model given below,
we pad the two sentences to have the same
length $s=\max(s_0,s_1)$. However, in the figures we show
different lengths because this gives a better intuition of
how the model works.

\begin{table}[tb]
 \setlength{\belowcaptionskip}{-10pt}
 \setlength{\abovecaptionskip}{0pt}
\begin{center}
\setlength{\tabcolsep}{2mm}
{\footnotesize
\begin{tabular}{l|r}
 symbol & description \\\hline
$s$, $s_0$, $s_1$ & sentence or sentence length\\
$v$ & word\\
$w$ & filter width\\
$d_i$ & dimensionality of input to layer $i+1$\\
$\mathbf{W}$ & weight matrix
\end{tabular}}
\end{center}
\caption{Notation}\label{tab:notation} 
\end{table}

\begin{figure}[t]
\setlength{\belowcaptionskip}{-10pt}
\setlength{\abovecaptionskip}{0pt}
\centering
\includegraphics[width=\colfigfactor\columnwidth]{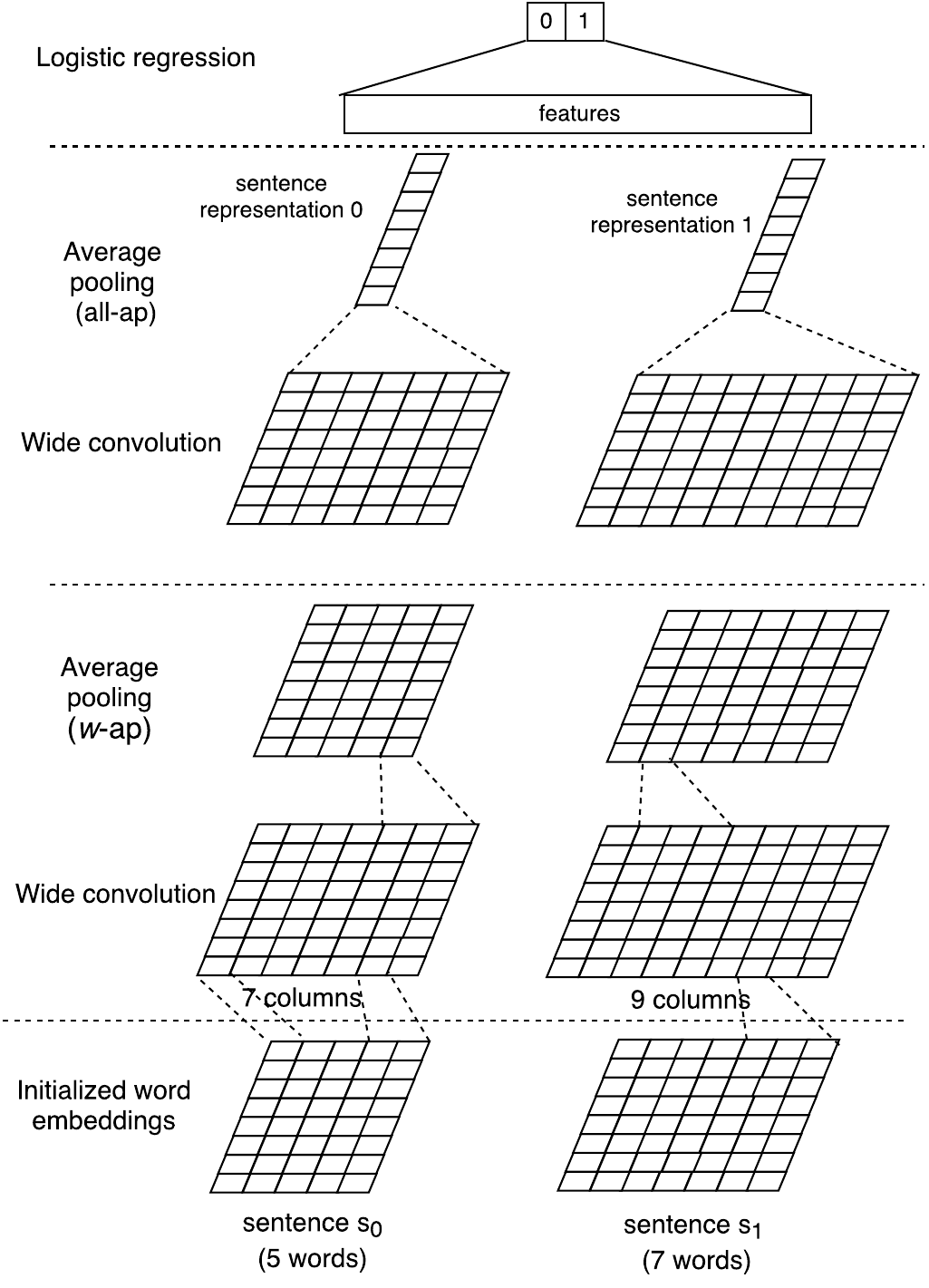}
\caption{BCNN: ABCNN without Attention} \label{fig:arc-0}
\end{figure}

We now describe the BCNN's  four types of layers: input, convolution,
average pooling  and output.

\enote{hs}{removed to save space

footnote{url{https://code.google.com/p/word2vec/}}

}

\textbf{Input layer.} In the example in the figure, the two
input sentences have 5 and 7 words, respectively.  Each word
is represented as a $d_0$-dimensional precomputed word2vec
\cite{mikolov2013distributed}
embedding, $d_0=300$.  
As a result, each sentence is
represented as a feature map of dimension $d_0 \times s$.

\textbf{Convolution layer.} Let $v_1,v_2,\ldots,v_s$ be the
words of a sentence and $\mathbf{c}_i\in\mathbb{R}^{w\cdot d_0}$,
$0< i <s+w$, the concatenated embeddings of
$v_{i-w+1},\ldots,v_{i}$ where embeddings for $v_j$  are set
to zero when $j<1$
or $j>s$.  We then generate the
representation $\mathbf{p}_i\in\mathbb{R}^{d_1}$ for the
\textit{phrase} $v_{i-w+1},\ldots,v_{i}$ using the
convolution weights $\mathbf{W}\in\mathbb{R}^{d_1\times
  wd_0}$ as follows:
\[\mathbf{p}_i=\mathrm{tanh}(\mathbf{W}\cdot\mathbf{c}_i+\mathbf{b})\]
where $\mathbf{b}\in\mathbb{R}^{d_1}$ is the bias.

\textbf{Average pooling layer.} Pooling (including min, max,
average pooling)
is commonly used to extract robust features from 
convolution.  In this paper, we introduce attention
weighting as an alternative, but use average pooling as a baseline as
follows.

For the output feature map of the last
convolution layer, we do column-wise averaging over \emph{all
columns}, denoted as \emph{all-ap}.
This generates 
a
representation vector for each of the two sentences, shown 
as the top ``Average pooling (\emph{all-ap})'' layer
below 
``Logistic regression'' in Figure \ref{fig:arc-0}.
These two vectors are the basis for the
sentence pair decision.

For the output feature map  of non-final convolution layers, we do column-wise averaging over
\emph{windows of $w$
consecutive columns}, denoted as  $w$-\emph{ap};
shown as 
the lower
``Average pooling ($w$-\emph{ap})'' layer in Figure \ref{fig:arc-0}.
For  filter width $w$,
a  convolution layer transforms an
input feature map of $s$ columns into a new feature map of
 $s+w-1$ columns; average pooling transforms this
back to $s$ columns. This architecture supports
stacking an arbitrary number of  convolution-pooling blocks
to extract increasingly abstract features. Input
features to the bottom layer are words, input features to
the next layer are short phrases and so on. Each level
generates more abstract features of higher granularity.



The last layer is an \textbf{output layer}, chosen according
to the task;
e.g., for 
binary classification tasks, this layer is
logistic regression
(see Figure \ref{fig:arc-0}).
Other types of output layers are introduced below.

We found that in most cases, performance is boosted if we
provide the output of \emph{all pooling layers} as input to the
output layer.  For each non-final average pooling layer, we
perform $w$-\emph{ap} (pooling over windows of $w$ columns) as
described above,
but we also perform \emph{all-ap}
(pooling over all columns)
and forward the result to the output
layer. This improves
performance because 
representations from
different layers cover the properties of the sentences at
different levels of abstraction and all of these
levels can be important for a particular sentence pair.

\section{ABCNN: Attention-Based BCNN}
\begin{figure*} \centering 
\subfigure[One block in ABCNN-1] { \figlabel{a} 
\includegraphics[width=1.8\columnwidth]{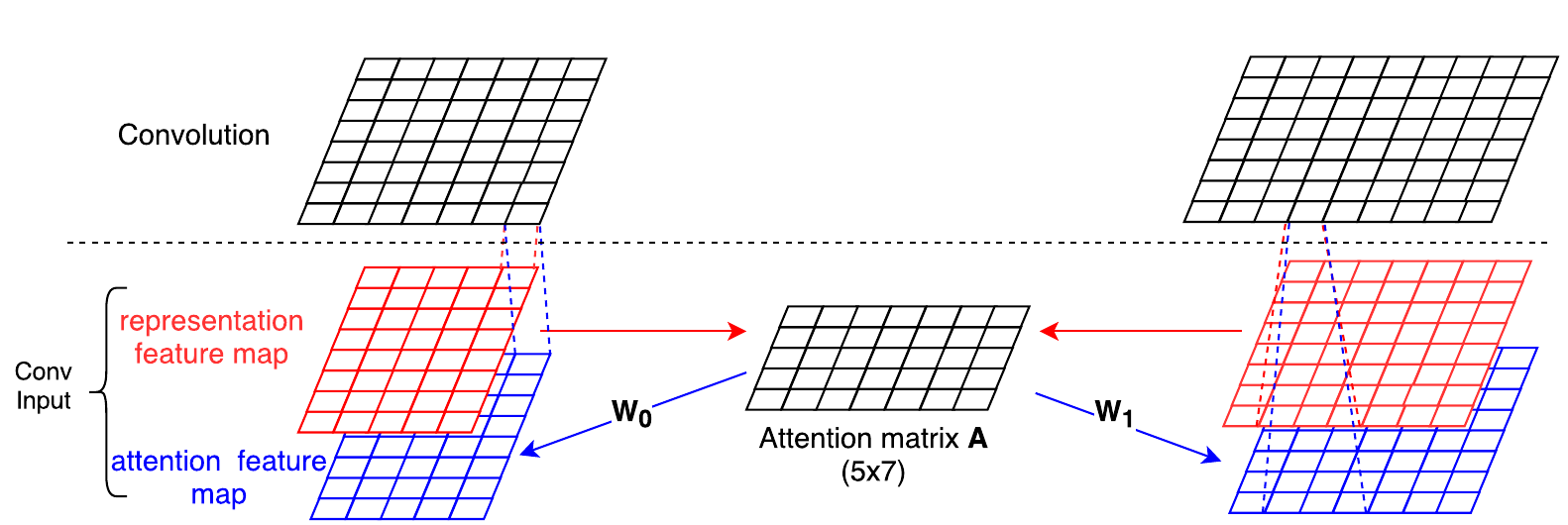} 
} 
\subfigure[One block in ABCNN-2] { \figlabel{b} 
\includegraphics[width=1.8\columnwidth]{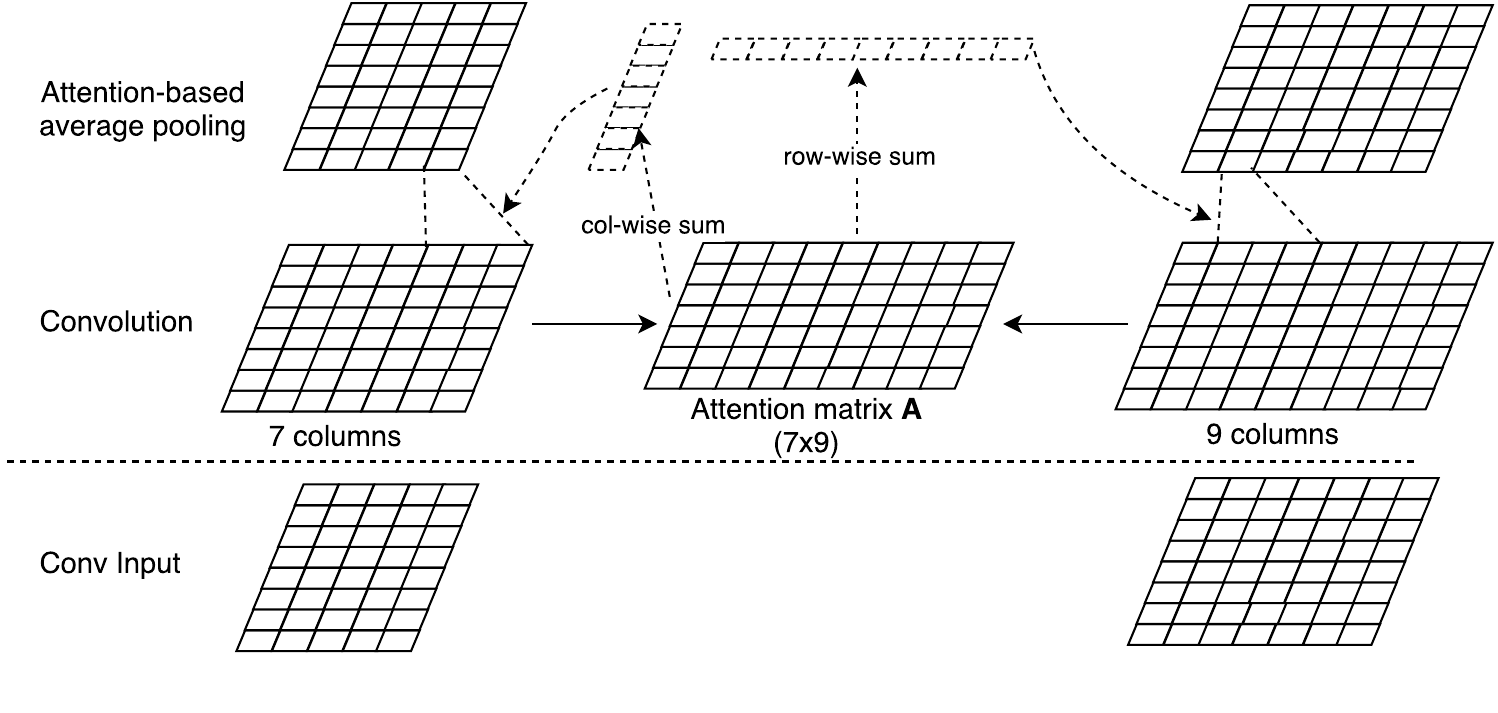} 
} 
\subfigure[One block in ABCNN-3] { \figlabel{c} 
\includegraphics[width=1.8\columnwidth]{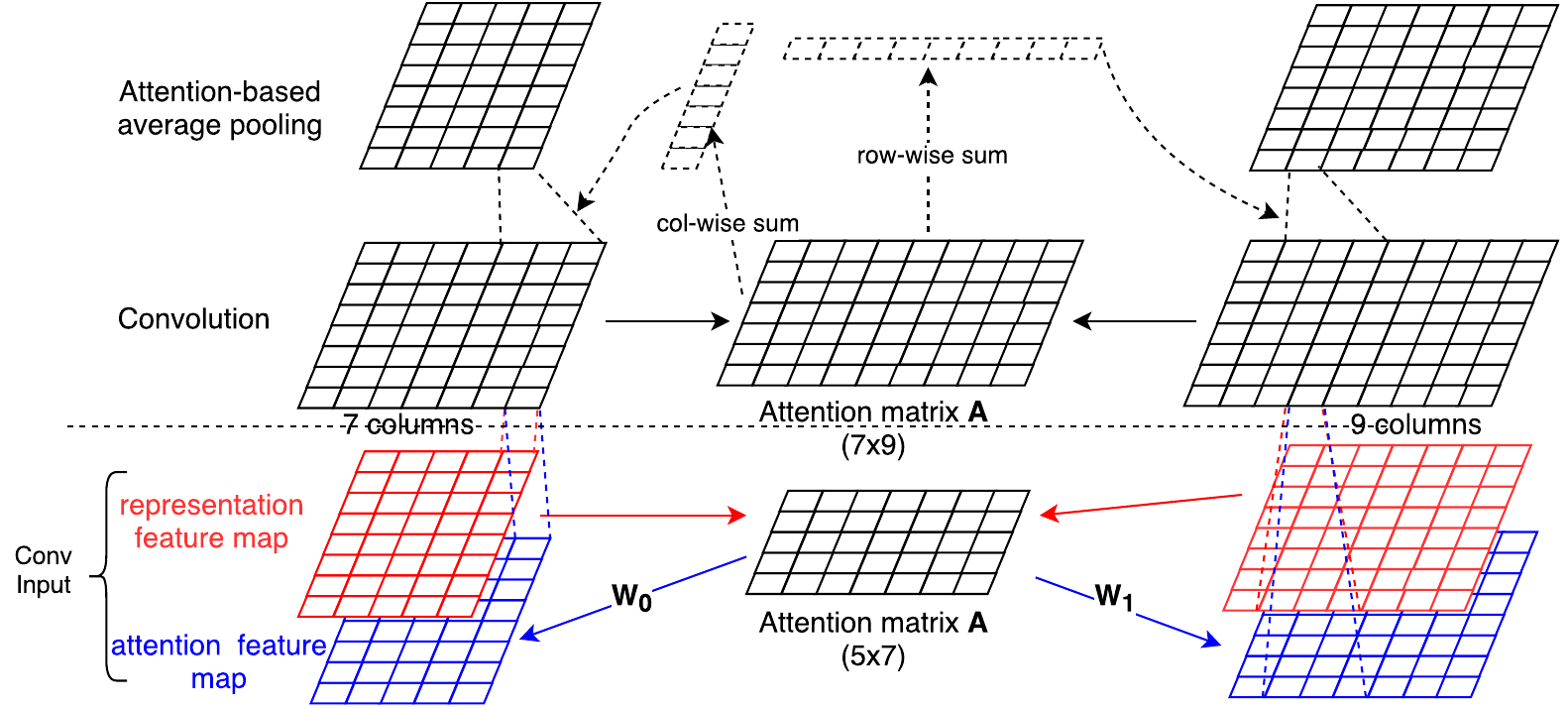} 
} 
\caption{Three ABCNN architectures} 
\label{fig:0} 
\end{figure*}

\seclabel{abcnn}
We now describe three 
architectures based on the BCNN, 
the ABCNN-1,
the ABCNN-2 and
the ABCNN-3, that each introduces an 
attention
mechanism for modeling sentence pairs; see
\figref{0}.

\textbf{ABCNN-1.}\label{sec:abcnn-1}
The ABCNN-1 (\figref{a}) employs an attention feature matrix
$\mathbf{A}$
to influence
convolution. Attention features are intended to weight those
units of $s_i$ more highly in convolution that are relevant
to a unit of $s_{1-i}$ \explainindexi;  we use the term ``unit'' here
to refer to words on the lowest level and to phrases on
higher levels of the network.
\figref{a} shows two \emph{unit representation feature maps}
in red: this part of the ABCNN-1 is the same as in
the BCNN (see \figref{arc-0}). Each column is
the representation of a unit,
a word  on the lowest level and a phrase
on higher levels.  We first describe
the attention feature
matrix $\mathbf{A}$ informally (layer ``Conv input'', middle
column, in \figref{a}).  $\mathbf{A}$  is generated by matching units
of the left representation feature map with units of the
right representation feature map such that the attention
values of row $i$ in $\mathbf{A}$ denote the attention
distribution of the $i$-th unit of $s_0$ with respect to
$s_1$, and the attention values of column $j$ in $\mathbf{A}$
denote the attention distribution of the $j$-th unit of
$s_1$ with respect to $s_0$.  $\mathbf{A}$ can be
viewed as a new feature map of $s_0$ (resp.\ $s_1$) in row
(resp.\ column) direction
because each
row (resp.\ column) is a new feature vector of a unit
in $s_0$ (resp.\ $s_1$).
Thus, it makes sense to combine this new feature map
with the representation feature maps and use both as
input to  the convolution operation. We achieve this by 
transforming
$\mathbf{A}$ into 
the two blue matrices in
Figure \ref{fig:a} that have the same format as the
representation feature maps.
As a result, the
new input of convolution  has two
feature maps for each sentence (shown in red and blue).
Our motivation is that the
attention
feature map will guide the convolution to learn
``counterpart-biased'' sentence representations.

More formally, 
let  $\mathbf{F}_{i,r}\in\mathbf{R}^{d\times
  s}$ be the \emph{representation feature map} 
of sentence
$i$ \explainindexi. Then we define the attention matrix
$\mathbf{A}\in\mathbf{R}^{s\times s}$ as follows:
\[\mathbf{A}_{i,j} =
\mbox{match-score}(\mathbf{F}_{0,r}[:,i],
\mathbf{F}_{1,r}[:,j]) \ \ \ \ \ \ \ \ \ \  \ (1)\]
The function match-score
can be defined in a variety of ways. We found that
$1/(1+|x-y|)$ works well where $| \cdot |$ is Euclidean distance.

Given attention matrix $\mathbf{A}$, we generate the
\emph{attention feature map} $\mathbf{F}_{i,a}$ for $s_i$ as follows:
\[\mathbf{F}_{0,a} = \mathbf{W}_0 \cdot \mathbf{A}^\top, 
\enspace\enspace\mathbf{F}_{1,a} = \mathbf{W}_1 \cdot \mathbf{A} \] 
The weight matrices
$\mathbf{W}_0\in\mathbf{R}^{d\times s}$,
$\mathbf{W}_1\in\mathbf{R}^{d\times s}$
 are parameters of the model to be learned in
 training.\footnote{The weights 
of the two  matrices are shared in our implementation to reduce the number
of parameters of the model.}

We stack the representation feature map $\mathbf{F}_{i,r}$
and the attention
feature map $\mathbf{F}_{i,a}$
as an order 3 tensor and  feed it
into convolution  to generate a higher-level representation feature
map for $s_i$ \explainindexi. In \figref{a}, $s_0$ has
5 units, $s_1$  has 7. The output of convolution
(shown in the top layer, filter width $w=3$) is a higher-level representation feature map
with 7 columns for $s_0$ and 
9 
columns for $s_1$.

\textbf{ABCNN-2.}\label{sec:abcnn-2}
The ABCNN-1 computes attention weights \emph{directly on the
input representation} with the aim of \emph{improving the features
computed by convolution}.  The ABCNN-2 (\figref{b}) instead computes attention
weights \emph{on the output of convolution} with the aim
of 
\emph{reweighting this convolution output}.
In the example shown in \figref{b}, the feature
maps output by convolution for $s_0$ and $s_1$ 
(layer marked ``Convolution'' in \figref{b})
have 7 and 9
columns, respectively; each column is the
representation of a unit.  The attention matrix $\mathbf{A}$
compares all units in $s_0$ with all units of $s_1$.  We sum
all attention values for a unit to derive a single attention
weight for that unit. This corresponds to summing all values
in a row of $\mathbf{A}$ for $s_0$ (``col-wise sum'', resulting in the column
vector of size 7 shown) and summing all values in a column
for $s_1$ (``row-wise sum'', resulting in the row vector of size 9 shown).

More formally, let $\mathbf{A}\in\mathbf{R}^{s\times s}$ be
the attention
matrix, $a_{0,j}=\sum\mathbf{A}[j,:]$ the attention weight
of unit $j$ in $s_0$,
$a_{1,j}=\sum\mathbf{A}[:,j]$ the attention weight of unit
$j$ in $s_1$ and
$\mathbf{F}^c_{i,r}\in\mathbf{R}^{d\times(s_i+w-1)}$ the
output of convolution
for $s_i$.
Then the $j$-th column of the new feature map
$\mathbf{F}^p_{i,r}$ generated by $w$-\emph{ap} is derived by:\\
\[\mathbf{F}^p_{i,r}[:,j]\!=\!\sum_{k=j:j+w}a_{i,k}\mathbf{F}^c_{i,r}[:,k], \enspace\enspace j=1 \ldots s_i\]
Note that $\mathbf{F}^p_{i,r}\in\mathbf{R}^{d\times s_i}$, i.e.,
ABCNN-2 pooling generates an output feature
map 
of the same size as the input feature map of convolution. This allows us
to stack multiple convolution-pooling
blocks to extract features of increasing abstraction.

There are three main differences between
the ABCNN-1 and the ABCNN-2.  (i)
Attention in the ABCNN-1 impacts \emph{convolution
indirectly} while attention in the ABCNN-2 influences \emph{pooling}
through \emph{direct} attention weighting. (ii) The ABCNN-1 
requires
the two matrices $\mathbf{W}_i$ to
convert
the attention matrix
into attention feature maps; and the input to
convolution has two times as many
feature maps. 
Thus, the ABCNN-1 has more parameters than the ABCNN-2 and
is more vulnerable to overfitting.
(iii) As pooling is performed
after convolution, 
pooling  handles larger-granularity units than
convolution; e.g., if the input to
convolution has word level granularity, then the input
to pooling has phrase level granularity, the
phrase size being equal to filter size $w$. Thus, the ABCNN-1
and the ABCNN-2 implement attention mechanisms for
linguistic units of different granularity. 
The complementarity of the ABCNN-1 and the ABCNN-2 motivates us to
propose the ABCNN-3, a third architecture that combines elements
of the two.



\textbf{ABCNN-3}
(Figure \ref{fig:c})
combines
the ABCNN-1 and the ABCNN-2 by stacking them; it
combines the strengths of the ABCNN-1 and -2 by
allowing the attention mechanism to operate (i)  both on the
convolution and on the pooling parts of a convolution-pooling
block and (ii) both on the input granularity and on the more
abstract output granularity. 


\section{Experiments}\label{sec:experiments} We test the
proposed
architectures on  three tasks: answer selection (AS),
paraphrase identification (PI) and textual entailment (TE).

\textbf{Common Training Setup.}\seclabel{cts}
Words are initialized by 300-dimensional word2vec
embeddings and not changed during training.  A single randomly initialized
embedding
is created for all
unknown words 
by uniform sampling from [-.01,.01].
We employ Adagrad \cite{duchi2011adaptive} and $L_2$
regularization.

\textbf{Network Configuration.}  Each network in the
experiments below consists of (i) an initialization block
$b_1$ that initializes words by word2vec embeddings, (ii) a
stack of $k-1$ convolution-pooling blocks $b_2,\ldots, b_k$,
computing increasingly abstract features, and (iii) one final
\emph{LR layer} (logistic regression layer) as shown in
\figref{arc-0}.

The input to the LR layer consists of $kn$ features -- each
block provides $n$ similarity scores, e.g., $n$ cosine
similarity scores.  \figref{arc-0} shows
the two sentence vectors output by the final block $b_k$ of
the stack (``sentence representation 0'', ``sentence
representation 1''); this is the basis of the last $n$ similarity scores.  As we explained in the final paragraph
of \secref{bcnn}, we perform \emph{all-ap} pooling for
\emph{all blocks}, not just for $b_k$.  Thus we get one
sentence representation each for $s_0$ and $s_1$ for each
block $b_1, \ldots, b_k$.  We compute $n$ similarity scores
for each block (based on the block's two sentence
representations). Thus, we compute a total of
$kn$ similarity scores and these scores are input to the LR layer.

Depending on the task,
we use different methods for computing the similarity score:
see below.


\textbf{Layerwise Training.}  In our training regime, we
first train a network consisting of just one convolution-pooling
block $b_2$. We then create a new network by adding a block $b_3$, initialize
its $b_2$ block with the previously learned weights for
$b_2$ and train $b_3$ keeping the
previously learned weights for $b_2$ fixed. We repeat this
procedure until all $k-1$ convolution-pooling blocks are trained.
We found that this training regime gives us good performance
and shortens training times considerably.  Since similarity
scores of lower blocks are kept unchanged once they have
been learned, this also has the nice effect that ``simple''
similarity scores (those based on surface features) are
learned first and subsequent training phases can focus on
complementary scores derived from more complex abstract
features.


\textbf{Classifier.} 
We found that performance increases if we
do not use the output of the LR layer as the final decision, but instead
train a linear SVM or a logistic regression with default parameters\footnote{
  \url{http://scikit-learn.org/stable/} for both.} directly on the input to
the LR layer (i.e., on the $kn$ similarity scores that are generated by the
$k$-block stack after
network training is completed). Direct training of SVMs/LR
seems to get closer to the global optimum than gradient
descent training of CNNs.

\tabref{hyper} shows 
hyperparameters, tuned on dev.

\def\hyperspace{0.075cm}

\begin{table}[t]
 \setlength{\belowcaptionskip}{-15pt}
 \setlength{\abovecaptionskip}{0pt}
\begin{center}
\setlength{\tabcolsep}{1mm}
{\footnotesize
\begin{tabular}{ll|l@{\hspace{\hyperspace}}l@{\hspace{\hyperspace}}l|l@{\hspace{\hyperspace}}l@{\hspace{\hyperspace}}l|l@{\hspace{\hyperspace}}l@{\hspace{\hyperspace}}l}
&&\multicolumn{3}{c|}{AS}&\multicolumn{3}{c|}{PI}&\multicolumn{3}{c}{TE}\\
&\rotatebox{90}{\#CL}&lr &$w$ &$L_2$ &lr &$w$ &$L_2$&lr &$w$ &$L_2$\\\hline
ABCNN-1 &1 &.08 &4 &.0004&.08 &3 &.0002 &.08 &3 &.0006 \\
ABCNN-1 &2 &.085 &4 &.0006 &.085 &3 &.0003 &.085 &3 &.0006 \\\hline
ABCNN-2 &1 &.05 &4 &.0003 &.085 &3 &.0001 &.09 &3 &.00065 \\
ABCNN-2 &2 &.06 &4 &.0006 &.085 &3 &.0001 &.085 &3 &.0007 \\\hline
ABCNN-3 &1 &.05 &4 &.0003 &.05 &3 &.0003 &.09 &3 &.0007 \\
ABCNN-3 &2 &.06 &4 &.0006 &.055 &3 &.0005 &.09 &3 &.0007 
\end{tabular}
}
\end{center}
\caption{Hyperparameters. lr: learning rate. \#CL: number
  convolution layers.  $w$: filter width. The number of
  convolution kernels $d_i$ ($i>0$) is
  50 throughout.}\label{tab:hyper} 
\end{table}

We use addition and LSTMs as  two \textbf{shared baselines} for all three tasks,
i.e., for AS, PI and TE. We now describe these two shared baselines.

(i) \textbf{Addition}. We sum up
word embeddings element-wise to form each sentence
representation. The classifier input is then the
concatenation of  the two sentence representations.
(ii) \textbf{A-LSTM}. Before
this work, most attention mechanisms in NLP were
implemented in recurrent neural networks for text generation
tasks such as machine translation
(e.g., \newcite{bahdanau2015neural}, \newcite{luong2015effective}).
\newcite{entail2016} present an attention-LSTM for
natural language inference. Since this model is the pioneering
attention based RNN system for sentence pair classification, we consider it as a baseline system
(``A-LSTM'') for all our three tasks. The A-LSTM has the same
configuration as our ABCNNs in terms of word
initialization (300-dimensional word2vec embeddings) and the
dimensionality of all hidden layers (50).
 
\subsection{Answer Selection}
\seclabel{as}
We use WikiQA,\footnote{\url{http://aka.ms/WikiQA}
  \cite{yang2015wikiqa}} an open domain
question-answer dataset. We use the subtask that assumes that
there is at least one correct answer for a question. 
The corresponding dataset consists of 20,360
question-candidate pairs in train, 1,130 pairs in dev and
2,352 pairs in test where we
adopt the standard setup 
of only
considering questions with correct answers in test.
Following
\newcite{yang2015wikiqa},
we  truncate answers to 40
tokens.

The task is to rank the candidate answers based on their
relatedness to the question. Evaluation measures are mean
average precision (MAP) and mean reciprocal rank (MRR).

\textbf{Task-Specific Setup.}
We use cosine similarity as the similarity score for AS. In addition, we use sentence lengths, \emph{WordCnt} (count
of the number
of non-stopwords in the question that also occur in the
answer) and
\emph{WgtWordCnt} (reweight the counts
by the IDF values of the question words). Thus, the final input to the LR layer has
size $k+4$: one cosine for each of the $k$ blocks and the four additional features.

We compare with seven \textbf{baselines}. The first three
are considered by \newcite{yang2015wikiqa}: (i) WordCnt;
(ii) WgtWordCnt; (iii) CNN-Cnt (the state-of-the-art system): combine CNN with (i) and
(ii). 
Apart from the baselines considered by 
\newcite{yang2015wikiqa}, we compare with two Addition
baselines and two LSTM baselines. Addition and A-LSTM are
the \emph{shared baselines} described before. We
also combine both with
the four extra features; this gives us two additional
baselines that we refer to as Addition(+) and A-LSTM(+).

\begin{table}[bt]
 \setlength{\belowcaptionskip}{-10pt}
 \setlength{\abovecaptionskip}{0pt}
\begin{center}
\setlength{\tabcolsep}{1mm}
{\footnotesize
\begin{tabular}{ll|ll}
&method &  MAP  & MRR \\ \hline
\multirow{7}{*}{\rotatebox{90}{Baselines}}&WordCnt & 0.4891 & 0.4924 \\
 & WgtWordCnt & 0.5099 & 0.5132\\
&CNN-Cnt & \underline{0.6520} & \underline{0.6652}\\
&Addition& 0.5021 & 0.5069\\
&Addition(+)& 0.5888 & 0.5929 \\
&A-LSTM& 0.5347 & 0.5483\\
&A-LSTM(+)& 0.6381 & 0.6537 \\\hline
\multirow{2}{*}{BCNN}&one-conv& 0.6629 & 0.6813\\
& two-conv & 0.6593 & 0.6738 \\\hline
\multirow{2}{*}{ABCNN-1}&one-conv& 0.6810$^*$ & 0.6979$^*$\\
& two-conv & 0.6855$^*$ & 0.7023$^*$ \\\hline
\multirow{2}{*}{ABCNN-2}&one-conv& 0.6885$^*$ & 0.7054$^*$\\
& two-conv & 0.6879$^*$ & 0.7068$^*$\\\hline
\multirow{2}{*}{ABCNN-3}&one-conv& 0.6914$^*$ & \textbf{0.7127}$^*$\\
& two-conv & \textbf{0.6921}$^*$ & 0.7108$^*$ 
\end{tabular}}
\end{center}
\caption{Results on
  WikiQA\tablabel{wikiqa}. Best result per column is bold. Significant
improvements over state-of-the-art baselines (underlined) are marked with $*$ ($t$-test, p $<$ .05).}
\end{table}

\textbf{Results.}
\tabref{wikiqa} 
shows performance of the baselines, of the BCNN and of the three
ABCNNs. For CNNs, we test one (one-conv) and two
(two-conv) convolution-pooling blocks.

The non-attention network BCNN already performs better than
the baselines. If we add attention mechanisms, then the
performance further improves by several points.  Comparing
the ABCNN-2 with the ABCNN-1, we find the ABCNN-2 is slightly better
even though the ABCNN-2 is the simpler architecture.  If we
combine the ABCNN-1 and the ABCNN-2 to form the ABCNN-3, we get
further improvement.\footnote{If we limit the input to the LR layer to the $k$ similarity
scores in
the ABCNN-3 (two-conv),
results are .660 (MAP) / .677 (MRR).}

 This can be explained by the ABCNN-3's
ability to take attention of finer-grained granularity
into consideration in each convolution-pooling block while
the ABCNN-1 and the ABCNN-2 consider attention only at convolution
input or only at pooling input, respectively.  We also find
that stacking two convolution-pooling blocks does not bring
consistent improvement and therefore do not test deeper
architectures.


\subsection{Paraphrase Identification}
We use the Microsoft Research Paraphrase  (MSRP) corpus
\cite{dolan2004unsupervised}. The training
set  contains 2753 true  / 1323 false
and
the test set 1147 true / 578 false
paraphrase pairs.
We
randomly select 400 pairs from  train and use them as 
dev; but we still report  results for
training on the entire training set. For each triple (label,
$s_0$, $s_1$) in the training set, we also add (label, $s_1$, $s_0$) to
the training set to make best use of the training data. Systems are evaluated 
by accuracy and $F_1$.

\textbf{Task-Specific Setup.}
In this task, we add the 15 MT features from
\cite{madnani2012re} and the lengths of the two sentences. In addition, we compute
ROUGE-1, ROUGE-2 and ROUGE-SU4 \cite{lin2004rouge},
which are scores measuring the match between the two
sentences on (i)
unigrams, (ii) bigrams and (iii) unigrams and skip-bigrams (maximum skip distance of four),
respectively. 
In this task, we found transforming
Euclidean distance into similarity score by
$1/(1+|x-y|)$ 
performs better than cosine
similarity. Additionally, 
we use dynamic pooling \cite{yinnaacl}
of the attention matrix $\mathbf{A}$ in Equation (1)
and forward pooled values of
all blocks to the classifier.
This
gives us  better performance than only forwarding
sentence-level matching features.

We compare our system 
with
representative DL approaches:
(i) A-LSTM; (ii) A-LSTM(+): A-LSTM plus handcrafted features; (iii) RAE
\cite{socher2011dynamic}, recursive autoencoder; (iv)
Bi-CNN-MI \cite{yinnaacl}, a bi-CNN architecture; and (v)
MPSSM-CNN \cite{he2015multi}, the state-of-the-art
NN system for PI, and the following four non-DL systems:
(vi) Addition;  (vii) Addition(+): Addition plus handcrafted features; (viii) MT
\cite{madnani2012re}, a system that combines machine translation metrics;\footnote{For better comparability of approaches in our experiments,
we use a simple SVM classifier, which  performs
slightly worse than \newcite{madnani2012re}'s more complex meta-classifier.}
(ix) MF-TF-KLD \cite{ji2013discriminative}, the state-of-the-art non-NN system.

\begin{table}[t]
\setlength{\belowcaptionskip}{-1pt}
\setlength{\abovecaptionskip}{0pt}
\begin{center}
\setlength{\tabcolsep}{1mm}
{\footnotesize
\begin{tabular}{ll|ll}
\multicolumn{2}{c|}{method} &  acc  & $F_1$ \\ \hline
\multirow{10}{*}{\rotatebox{90}{Baselines}}&majority voting & 66.5 & 79.9 \\
 & RAE & 76.8 & 83.6\\
&Bi-CNN-MI & 78.4 & 84.6\\
&MPSSM-CNN & \underline{78.6}  & \underline{84.7}\\ 
& MT & 76.8 & 83.8\\
&MF-TF-KLD & \underline{78.6} & 84.6\\
&Addition & 70.8 & 80.9\\
& Addition (+) & 77.3 & 84.1 \\
&A-LSTM & 69.5 & 80.1\\
& A-LSTM (+) & 77.1 & 84.0 \\\hline
\multirow{2}{*}{BCNN}&one-conv& 78.1 & 84.1\\
& two-conv & 78.3 & 84.3 \\\hline
\multirow{2}{*}{ABCNN-1}&one-conv& 78.5 & 84.5\\
& two-conv & 78.5 & 84.6 \\\hline
\multirow{2}{*}{ABCNN-2}&one-conv& 78.6 & 84.7\\
& two-conv & 78.8 & 84.7\\\hline
\multirow{2}{*}{ABCNN-3}&one-conv& 78.8 & \textbf{84.8}\\
& two-conv & \textbf{78.9} & \textbf{84.8} 
\end{tabular}}
\end{center}
\caption{Results for PI on MSRP}\label{tab:msrp} 
\end{table}

\textbf{Results.}
\tabref{msrp} shows that the BCNN is slightly
worse than the state-of-the-art whereas  the ABCNN-1 roughly
matches it.
The ABCNN-2 is slightly above the
state-of-the-art. The ABCNN-3 outperforms
the state-of-the-art  in accuracy and
$F_1$.\footnote{Improvement of .3 (acc) and .1 ($F_1$) over
state-of-the-art is not significant.
The ABCNN-3 (two-conv) without  ``linguistic''
features (i.e., MT and ROUGE)
achieves 75.1/82.7.}
Two
convolution layers only bring small improvements over one.

\subsection{Textual Entailment}
SemEval 2014 Task 1 \cite{marelli2014semeval} evaluates
system predictions of textual entailment (TE) relations on
sentence pairs from the SICK dataset
\cite{marelli2014sick}. The three classes are
entailment, contradiction and neutral.  The
sizes of SICK train, dev and test sets are 4439, 495 and
4906 pairs, respectively.  We call this dataset ORIG.

We also create NONOVER, a copy of ORIG in which \emph{words
occurring in both sentences are removed}.  A sentence
in NONOVER is denoted by the special token $<$empty$>$
if all words are removed. \tabref{te} shows three pairs from
ORIG  and their transformation in NONOVER.  We
observe that focusing on the non-overlapping parts provides
clearer hints for TE than ORIG.  
In this task, we run two copies of each network, one for
ORIG, one for NONOVER; these two networks have a single
common LR layer.

Like \newcite{lai2014illinois}, we train our final
system (after fixing hyperparameters) on train and dev
(4934 pairs).  Eval measure is accuracy.

\textbf{Task-Specific Setup.}
We found that for this task forwarding two similarity scores
from each block (instead of just one) is helpful. We use
cosine similarity and Euclidean distance.  As we did for paraphrase
identification, we add the 15 MT features for each sentence
pair for this task as well; our motivation is that entailed
sentences resemble paraphrases more than contradictory
sentences do.
 
\begin{table}[t]
\setlength{\belowcaptionskip}{-10pt}
\setlength{\abovecaptionskip}{5pt}
\setlength\tabcolsep{3pt}
\centering
{\footnotesize
\begin{tabular}{l|ll}
\rotatebox{90}{} & ORIG  &NONOVER\\\hline
\multirow{3}{*}{0} & children  in red shirts are   & children red shirts\\
&playing in the leaves& playing  \\\cdashline{2-3}
& three kids  are sitting in the leaves  & three kids sitting \\\hline
\multirow{2}{*}{1} & three boys are jumping in the leaves & boys \\\cdashline{2-3}
& three kids are jumping in the leaves & kids \\\hline
\multirow{2}{*}{2} & a man is jumping into an empty pool&an empty\\\cdashline{2-3}
& a man is jumping into a full pool & a full 
\end{tabular}
}
\caption{SICK data: Converting the original sentences (ORIG)
  into the NONOVER format}\label{tab:te}
\end{table}

We use the following linguistic features.
\textbf{Negation}
is important  for detecting contradiction.
Feature \textsc{neg}
is set to $1$ if either sentence contains
``no'', ``not'', ``nobody'', ``isn't'' and  to $0$ otherwise.
Following
\newcite{lai2014illinois}, we  use WordNet \cite{miller1995wordnet} to
detect 
\textbf{nyms}: 
synonyms, hypernyms and antonyms in the pairs. But
we do this on NONOVER (not on ORIG) to focus on what is
critical for TE.
Specifically, feature \textsc{syn} is the number of
word pairs in $s_0$ and $s_1$ that are synonyms. \textsc{hyp0}
(resp.\ \textsc{hyp1}) is
the number of words in $s_0$ (resp.\ $s_1$) that have a
hypernym in $s_1$ (resp.\ $s_0$).
In addition, we collect all
\emph{potential antonym pairs} (PAP) in NONOVER. We
identify the matched chunks that occur in
\emph{contradictory} and
\emph{neutral}, but not in \textit{entailed} pairs. We
exclude synonyms and hypernyms and apply a frequency filter
of $n = 2$. In contrast to \newcite{lai2014illinois}, we
constrain the PAP pairs to cosine similarity above 0.4 in
word2vec embedding space as this   discards many
noise pairs. Feature \textsc{ant} is the number of matched
PAP antonyms in a sentence pair.
As before we use sentence \textbf{lengths}, both for ORIG
(\textsc{len0o}: length $s_0$,
\textsc{len1o}: length $s_1$) and
for 
NONOVER
(\textsc{len0n}: length $s_0$,
\textsc{len1n}: length $s_1$).

On the whole, we have 24 extra features: 15 MT metrics,
\textsc{neg}, \textsc{syn}, \textsc{hyp0}, \textsc{hyp1}, \textsc{ant},
\textsc{len0o},
\textsc{len1o},
\textsc{len0n} and
\textsc{len1n}.

Apart from the Addition and LSTM baselines, we further
compare with the top-3 systems in SemEval and TrRNTN
\cite{bowman2015recursive},  a recursive neural network
developed for this SICK task.

\textbf{Results.}
\tabref{sickresult} shows that our CNNs outperform 
A-LSTM (with or without linguistic features added) and
the top three
SemEval systems.
Comparing
ABCNNs with the BCNN, attention mechanisms consistently improve
performance. The ABCNN-1 has performance
comparable to the ABCNN-2 while the ABCNN-3 is better still: a boost of
1.6
points compared to the previous state of the art.\footnote{If
we run the ABCNN-3 (two-conv) without the 24 linguistic
features, performance is 84.6.}


\begin{table}[!t]
\setlength{\belowcaptionskip}{-15pt}
\setlength{\abovecaptionskip}{0pt}
\begin{center}
\setlength{\tabcolsep}{1mm}
{\footnotesize
\begin{tabular}{ll|l}
\multicolumn{2}{c|}{method} &  acc  \\ \hline\hline
\multirow{3}{*}{\rotatebox{90}{\footnotesize \begin{tabular}{c}Sem-\\ Eval\\ Top3\end{tabular}}}&\cite{jimenez2014unal}
& 83.1 \\
 & \cite{zhao2014ecnu} & 83.6 \\
&\cite{lai2014illinois}  & \underline{84.6} \\\hdashline
TrRNTN & \cite{bowman2015recursive}& 76.9\\\hdashline
\multirow{2}{*}{Addition}&no features& 73.1\\
& plus features &  79.4 \\\hdashline
\multirow{2}{*}{A-LSTM}&no features& 78.0\\
& plus features &  81.7 \\\hline\hline
\multirow{2}{*}{BCNN}&one-conv& 84.8 \\
& two-conv &  85.0 \\\hline
\multirow{2}{*}{ABCNN-1}&one-conv& 85.6 \\
& two-conv &  85.8 \\\hline
\multirow{2}{*}{ABCNN-2}&one-conv& 85.7  \\
& two-conv & 85.8 \\\hline
\multirow{2}{*}{ABCNN-3}&one-conv& 86.0$^*$  \\
& two-conv & \textbf{86.2}$^*$   
\end{tabular}}
\end{center}
\caption{Results on SICK. Significant
improvements over \protect\cite{lai2014illinois} are marked
with $*$ (test of equal proportions, p $<$ .05). }\label{tab:sickresult} 
\end{table}

\def\vizfactor{\colfigfactor}

\begin{figure}[!ht] \centering 
\includegraphics[width=\vizfactor\columnwidth]{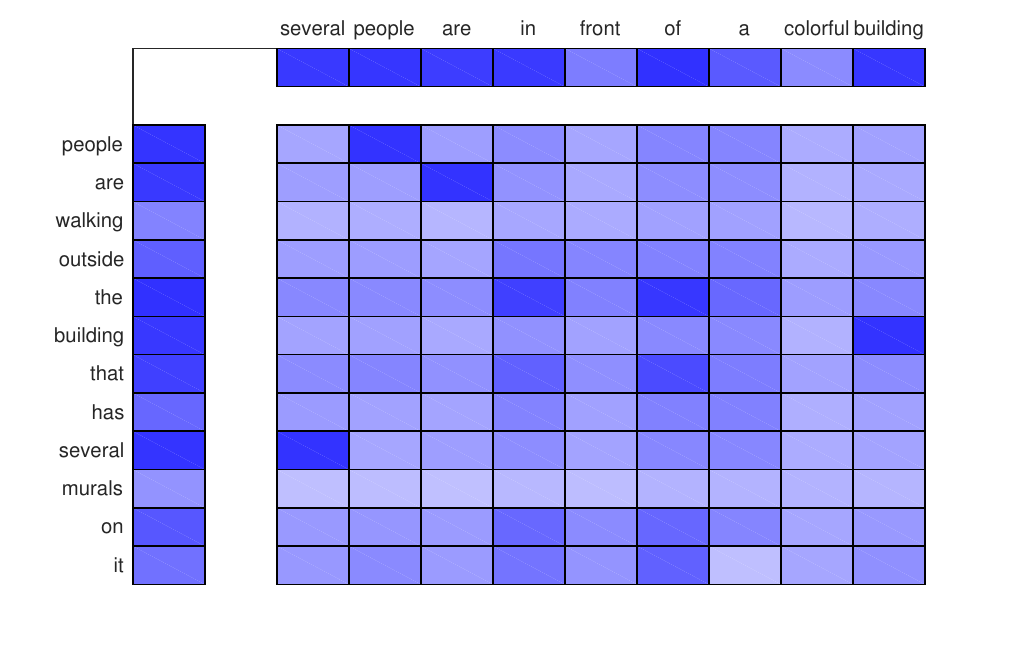}
\includegraphics[width=\vizfactor\columnwidth]{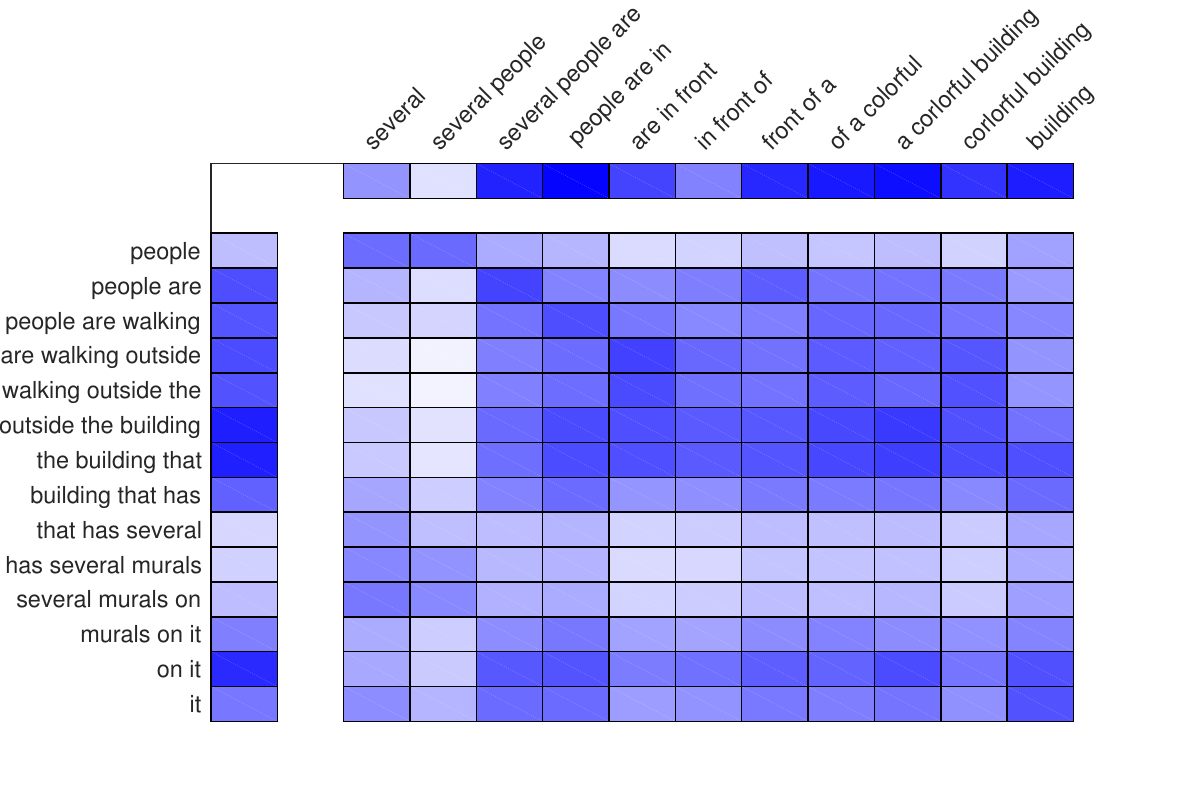}
\includegraphics[width=\vizfactor\columnwidth]{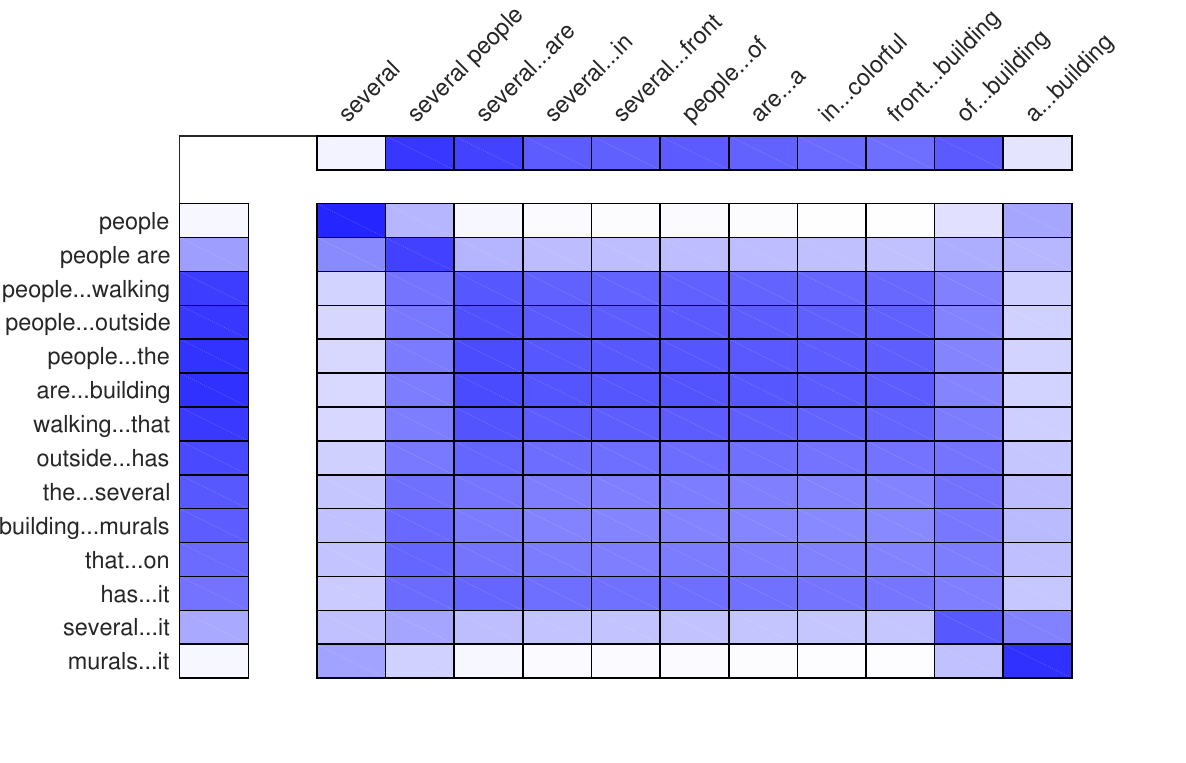}
\caption{Attention visualization for TE. Top:
  unigrams, $b_1$. Middle: conv1, $b_2$. Bottom: conv2, $b_3$.
 \figlabel{visual}}
\end{figure}

\textbf{Visual Analysis.}
\figref{visual} visualizes the attention matrices for
one TE sentence pair in the ABCNN-2 for blocks
$b_1$ (unigrams), $b_2$ (first convolutional layer) and $b_3$
(second convolutional layer). Darker shades of blue indicate
stronger attention values.

In \figref{visual} (top), each word
corresponds to exactly one row or column. We can see that
words in $s_i$ with semantic equivalents in $s_{1-i}$ get
high attention while words without semantic equivalents get
low attention, e.g., ``walking'' and ``murals'' in $s_0$ and
``front'' and ``colorful'' in $s_1$. This behavior seems
reasonable for the unigram level.

Rows/columns of the attention matrix in \figref{visual}
(middle) correspond to phrases of length three since
filter width $w=3$.  
High attention values generally correlate with close
semantic correspondence:
the phrase ``people are'' in $s_0$
matches ``several people are'' in $s_1$; both ``are walking
outside'' and ``walking outside the'' in $s_0$ match ``are
in front'' in $s_1$; ``the building that'' in $s_0$ matches
``a colorful building'' in $s_1$. More interestingly, looking at the bottom
right corner, both ``on it'' and ``it'' in $s_0$ match
``building'' in $s_1$; this indicates that ABCNNs are able to
detect some coreference across sentences. ``building'' in
$s_1$ has two places in which higher attentions appear, one
is with ``it'' in $s_0$, the other is with ``the building
that'' in $s_0$. This may indicate that ABCNNs recognize  that ``building'' in $s_1$
and ``the building that'' / ``it'' in $s_0$
refer to the same object. Hence, coreference resolution across sentences
as well as within a sentence both are detected. For the
attention vectors on the left and the top, we can see that
attention has focused on the key parts: ``people are walking
outside the building that'' in $s_0$, ``several people are
in'' and ``of a colorful building'' in $s_1$.

Rows/columns of the attention matrix in
\figref{visual} (bottom, second layer of convolution) correspond to phrases of length 5 since
filter width $w=3$ in both convolution layers ($5=1+2*(3-1)$).
We use ``$\ldots$'' to denote words in the middle if a phrase
like ``several...front'' has more than two words. We can
see that attention distribution in the matrix has focused on
some local regions. As granularity of phrases is larger, it
makes sense that the attention values are smoother. But
we still can find some interesting clues: at the two ends of
the main diagonal, higher attentions hint that the first
part of $s_0$ matches well with the first part of $s_1$;
``several murals on it'' in $s_0$ matches well with ``of a
colorful building'' in $s_1$, which satisfies the intuition
that these two phrases are crucial for making a decision on TE in this case.
This again shows the potential strength of our
system in figuring out which parts of the two sentences
refer to the same object. In addition, in the
central part of the matrix, we can see that the long phrase
``people are walking outside the building'' in $s_0$ matches
well with the long phrase ``are in front of a colorful
building'' in $s_1$.

\section{Summary}\label{sec:sum}
We presented three mechanisms to integrate
attention into CNNs for general
sentence pair modeling tasks.

Our experiments on AS, PI and TE
show that attention-based CNNs perform better than CNNs
without attention mechanisms.  The ABCNN-2 generally outperforms
the ABCNN-1 and the ABCNN-3 surpasses both. 

In all tasks, we did not
find any big improvement of two layers of convolution over
one layer. This is
probably due to the limited size of training data. We expect
that, as larger training sets become available, deep ABCNNs
will show even better performance. 

In addition, linguistic features contribute in all three
tasks: improvements by 0.0321 (MAP) and 0.0338 (MRR) for AS,
improvements by 3.8 (acc) and 2.1 ($F_1$) for PI and
an improvement by 1.6 (acc) for TE. But our ABCNNs can still
reach or surpass state-of-the-art even without those
features in AS and TE tasks. This indicates that ABCNNs are
generally strong NN systems.

Attention-based LSTMs are especially successful in tasks
with a strong \emph{generation component}
like machine translation (discussed in Sec.\ 2). CNNs have not been used for this type of task.
This is an  interesting area of 
future work for attention-based CNNs.

\subsection*{Acknowledgments}
We gratefully acknowledge the support of Deutsche
Forschungsgemeinschaft (DFG): grant SCHU 2246/8-2.

We would like to thank the anonymous reviewers for their
helpful comments.

\bibliographystyle{acl2012}
\bibliography{acl2012}

\end{document}

%% file: new1.tex
\textbf{Non-DL on Sentence Pair Modeling.}
  Sentence
pair modeling has attracted lots of attention in the past
decades. Many tasks can be reduced to a semantic text
matching problem. In this paper, we adopt the arguments by
  \newcite{yih2013question} who argue against shallow
  approaches as well as against
semantic text matching approaches that can be
computationally expensive:
\begin{quote}
Due to the variety of word choices and
inherent ambiguities in natural language, bag-of-word
approaches with simple surface-form word matching tend to
produce brittle results with poor prediction accuracy
\cite{bilotti2007structured}. As a result, researchers
put more emphasis on exploiting syntactic and
semantic structure. Representative examples
include methods based on deeper semantic analysis
\cite{shen2007using,moldovan2007cogex}, tree
edit-distance \cite{punyakanok2004mapping,heilman2010tree}
and quasi-synchronous grammars \cite{wang2007jeopardy} that
match the dependency parse trees of the two sentences.
\end{quote}Instead
of focusing on the high-level semantic representation,
\newcite{yih2013question} turn their attention to improving the
shallow semantic component, lexical semantics, by
performing
semantic matching based on 
a latent word-alignment structure 
(cf.\ \newcite{chang2010discriminative}). 
\newcite{lai2014illinois} explore
finer-grained word
overlap and alignment between two sentences using
negation, hypernym,
synonym and antonym relations.
\newcite{yao2013semi} extend word-to-word
alignment to phrase-to-phrase alignment by a semi-Markov
CRF. However, such approaches often require more
computational resources. In addition, employing
syntactic or semantic parsers -- which 
produce errors on many sentences --
to find the
best match between the structured representations of  two sentences
is not trivial.

%% file: new2.tex
\newcite{mnih2014recurrent} apply attention in recurrent neural
networks (RNNs) to extract ``information from an image or
video by adaptively selecting a sequence of regions or
locations and only processing the selected regions at high
resolution.''